# An Artificial Intelligence-based Framework to Achieve the Sustainable Development Goals in the Context of Bangladesh

Md. Tarek Hasan[1], Mohammad Nazmush Shamael[1], Arifa Akter[1], Rokibul Islam[1], Md. Saddam Hossain Mukta[1], and Salekul Islam[1]

**Abstract**

Sustainable development is a framework for achieving human development goals. It provides natural systems' ability to deliver natural resources and ecosystem services. Sustainable development is crucial for the economy and society. Artificial intelligence (AI) has attracted increasing attention in recent years, with the potential to influence many domains positively. AI is a commonly employed component in the quest for long-term sustainability. In this study, we explore the impact of AI on three pillars of sustainable development: society, environment, and economy, as well as numerous case studies from which we may deduce the impact of AI in a variety of areas, i.e., agriculture, classifying waste, smart water management, and Heating, Ventilation, and Air Conditioning (HVAC) systems. Furthermore, we present AI-based strategies for achieving Sustainable Development Goals (SDGs), which are effective for developing countries like Bangladesh. The framework that we propose may reduce the negative impact of AI and promote the proactiveness of this technology.

## 1. Introduction

Sustainable development is defined as the growth of society, the environment, and the economy in such a way that it adds value to future generations. The term "sustainable development" refers to the development that fulfills current demands without jeopardizing the ability of future generations to satisfy their own needs [1]. To achieve sustainability, the viability of the economy, protectivity of the environment, and equitability of society need to be ensured. A developing country like Bangladesh needs a well-thought-out strategy to attain long-term development. We are currently extremely linked to smart technologies like Artificial Intelligence (AI), and we can make a difference by utilizing the beneficial consequences of these technologies.

AI can be used in every industry to open up new possibilities and increase efficiency. In the scenario of Bangladesh, AI has the potential to significantly cut human labor while ensuring task completion. We need to use technology as much as possible in diverse areas of our nation to accomplish the SDGs, but we must remember the three pillars of sustainable development. We cannot neglect any of the pillars in order to achieve sustainability, thus we should also analyze the negative impacts of AI. The use of AI technology can enrich the economic growth of a country as it ensures productivity.

Shamsuzzaman et al. [2] outline the several factors that have led to the inadequate fisheries regulatory framework of Bangladesh. The research looks into how the legal structure of the Bangladeshi fishing sector might be improved. According to their study, a reformed legal framework they propose in Bangladesh might help the country achieve SDG #14: Life Below Water. Islam et al. [3] develop a thorough conservation strategy for the coastal and marine ecosystem of Bangladesh. They also focus on SDG #14 and suggest a framework for playing

---

[1] Department of Computer Science and Engineering, United International University

an incremental role in boosting blue growth and achieving SDGs. Sourav et al. [4] propose a framework for developing a smart city for a densely populated country like Bangladesh to achieve sustainable development. The findings of their research provide a clear image of the key features of smart cities and the elements that determine their sustainability. Roy et. al [5] state that altering the framework around this nexus of sustainable development initiatives may boost the productive basis of the economy and provide the impetus for the people of Bangladesh to achieve equal human welfare. Although the previous studies are well-defined, employing AI for sustainable development helps us achieve our objective more quickly, which is an urgent need in our country.

After reviewing related articles, we have noticed that there is no suggested framework for acquiring SDG by executing AI in the case of Bangladesh. There are a few works available on the impact of AI on achieving sustainable development goals. So, AI-based frameworks will create a new extent in attaining sustainable development goals.

In this study, we analyze the influence of AI on three pillars of sustainable development: society, environment, and economics. Furthermore, we propose an AI-based framework to accomplish sustainable growth in four distinct sectors: agriculture, waste management, water management, and heating, ventilation, and air conditioning (HVAC) systems. In addition, we explore the negative effects of AI and possible solutions to mitigate these effects.

## 2. AI and sustainable development

The introduction of AI has ushered in a new era of computing where computers can do tasks that usually require human intelligence to complete. At the core, AI is a field that merges computer science, mathematics, and data science to enable problem-solving. Machine learning [13], and Deep Learning [14] are subfields of AI that are often discussed alongside AI. These fields have been shaping an increasing set of sectors. With its ever-increasing reach, AI can have a significant impact on sustainable development. From Google's use of AI to reduce the energy cost of cooling of its data center by 40% [15] to Xcel Energy's use of AI to predict power consumption patterns to burn coal more efficiently, AI has already started to show its usefulness in sustainable development.

Sustainable development encompasses the three pillars: social, environmental, and economic. In order to achieve sustainable development, we want to achieve a balance of economic growth, environmental sustainability, and social inclusiveness. In this section, we discuss how AI can impact these three pillars of sustainable development.

### 2.1 Social Pillar
AI can be used to distribute food, health, water, and energy resources more efficiently. More efficient smart cities based on AI can be low carbon-emitting because of their efficient use of their resources. Technologies like interconnected autonomous cars can reduce traffic loads and thus reduce pollution caused by traffic. AI can help us integrate renewable energy sources like wind and solar energy with our existing energy sources by predicting when to reduce burning fuel based on the weather. It creates new job opportunities and reduces overall labor-intensive work.

### 2.2 Environmental Pillar
AI enables us to fight climate change by enabling us to better utilize our resources. We can utilize AI to monitor and predict energy consumption to create more efficient power generation systems. AI may cause over-exploitation of natural resources which damages the environment greatly. As AI integration is not as prominent now, it is still too early to be sure that AI can have a net positive impact on our environment. It can help us improve and conserve biodiversity by monitoring and predicting environmental changes. These systems help us with environmental planning, decision making, and environmental resource management.

**2.3 Economic Pillar**

The introduction of AI can have a massive impact on the economy of a nation. Automation using AI can reduce manual work and take over many dangerous tasks from humans. With the integration of AI, we can design and create more fast and sustainable infrastructures. This can lead to more responsible consumption and production. AI technologies can be very costly to develop. This creates more differences between the economies of low to mid-income countries and high-income countries. With AI replacing old manual jobs with jobs that require more technical knowledge, it has a heavy bias towards the educated.

**3. Case Studies**

For the above-mentioned four domains: agriculture, waste management, clean water, and climate change, we illustrate the relationship between AI and sustainable development; our case studies reflect similar fields. There are numerous additional sectors where AI has been effectively used for sustainable development, but for the sake of space, we confine our study to these four.

**3.1 Agriculture and AI: a case of sustainable agriculture**
**The See & Spray system:** The system from Blue River Technology allows farmers to spray herbicide exclusively to weeds, using less than a tenth of the herbicide used in traditional weed treatment. See and Spray has developed the first equipment in the world that allows farmers to improve every plant in their field using machine learning and robotics [6][2].

**The Sowing Application:** AI is being used in agriculture by Microsoft and ICRISAT to increase agricultural yields[3]. The results reveal a 30 percent greater average yield per hectare following the introduction of the pilot in June 2016, which tested a new Sowing Application for farmers paired with a Personalized Village Advisory Dashboard for the Indian state of Andhra Pradesh. The Sowing App was created to assist farmers in achieving optimal harvests by recommending the ideal time to plant-based on weather, soil, and other factors. The pilot was carried out in Kurnool's Devanakonda Mandal, and the advice solely applied to the groundnut crop.

**3.2 AI and the Global Water Crisis: a case of smart water management**
An Artificial Neural Network (ANN) based water need forecasting system was developed by the Spanish Ministry of Economy says [19]. The system connects ANN with the Bayesian framework and Genetic Algorithms (GA) to provide short-term day-to-day forecasting of water irrigation requirements. The method has enhanced the prediction exactness by 11% compared to non-AI-based techniques.

Water contamination issues can be detected by AI. By using trained models to identify dangerous particles and bacteria. Broadcasting devices that observe water for issues help municipalities to notice contamination as soon as possible.

**3.3 An AI Approach for Waste Management and Recycling: a case of classifying waste**
**Clean Robotics** - TrashBot™ has developed an automated system that detects and separates landfills from recyclables using robotics, computer vision, and AI. It performs this more accurately than humans, collects high-quality trash data, and alerts personnel when the bin is full. Individual units can learn from the global TrashBot fleet and become more intelligent over time thanks to cloud connectivity. A monitor for business communications, education, and advertising are also included. According to CleanRobotics claim, with the help of AI TrashBot can separate landfills vs recyclables with more than 90% of accuracy where traditional mechanisms without AI are about 30% accurate.

---

[2] See & Spray - the world's first smart sprayer
[3] Microsoft and ICRISAT - Bringing Artificial Intelligence to agriculture to boost crop yield

A piece of trash is tossed into the TrashBot, which is subsequently covered by a little door. A camera is hidden behind that door, analyzing the sort of garbage. In addition, the object is weighed on a Teflon-coated plastic shelf that drains any moisture. CleanRobotics software assesses whether the item is going to a landfill or a recycling center, and then guides it to the proper container below. On technical terms, TrashBot does this by using image processing, computer vision, and AI. CleanRobotics additionally offers a monthly software subscription that allows more waste management facilities.

### 3.4 AI in making efficient HVAC systems

One of the largest sources of energy consumption is the heating, ventilation, and air conditioning system also known as HVAC systems. While there have been many advancements in the field of HVAC systems to make them more energy efficient, the field of HVAC systems have reached maturity and not much can be done to the hardware to make them more efficient. In recent years, the integration of AI in HVAC systems has shown promising improvement in energy consumption of the HVAC systems. DeepMind's machine learning model has proven to be highly efficient at reducing the energy consumption for google's data centers[15]. By integrating the ML-based cooling system into Google's data center, the energy consumption related to cooling is reduced by up to 40%. Given how advanced and complicated google's data centers are and how widely they are used, this is a very significant improvement. One of the primary sources of energy consumption in a data center comes from cooling. The computers in the data center generate a lot of heat. DeepMind took the historical data collected by thousands of sensors within Google's data center to train the Deep Neural Network model. These sensors monitored parameters like power consumption, temperature, pump speed, set points etc. The model was trained to reduce the average future Power Usage Effectiveness or PUE of the data center. The model can ensure that the data center won't go beyond operating constraints by predicting how much resources will be needed in the future based on the pressure and temperature of the data center. Major breakthroughs like this show us the huge potential ML has at reducing energy consumption and help us get one step closer to combating climate change.

### 4. AI-based Framework for Sustainable Development

In the context of Bangladesh, we propose a framework for the four major systems to accomplish the SDGs. We can achieve several SDGs by assuring the sustainability of agriculture, water management, waste management, and HVAC systems, which can lead to the sustainability of a developing country like Bangladesh.

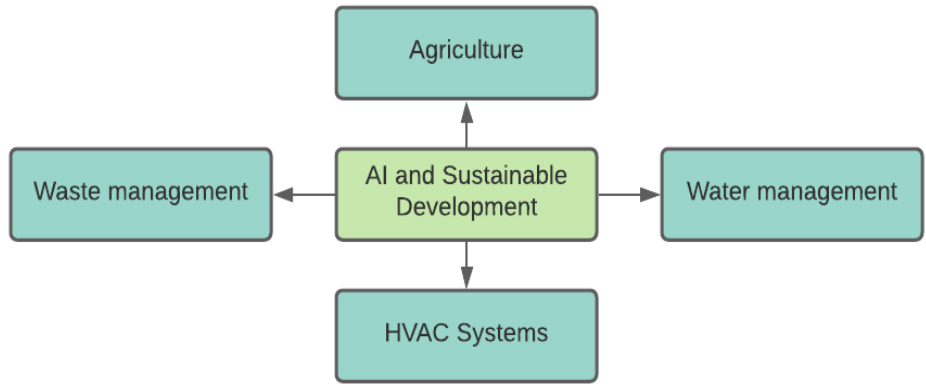

**Fig. 1:** Four major systems to achieve the SDGs in the context of Bangladesh

## 4.1 Decision-Making Support System for Agriculture

Agriculture must be prioritized in order to achieve SDG #2: Zero Hunger, and SDG #8: Decent Work and Economic Growth. Agriculture is an important component of economic development. The majority of the farmers of our country struggle to make informed decisions based on crop, soil, and weather conditions. AI makes decision-making easier by making it quicker and smarter. Using the benefits of AI, we propose a decision-making support system for farmers that will assist them in taking appropriate actions depending on many parameters.

We can employ a drone to take the image of crops, which is usually applied for this sort of task. Following the capture of photographs, we can send the data to an AI system, which will determine the crop condition. YOLO, invented by Redmon et al. [7], is one of the most widely used algorithms for this purpose. Currently, the state-of-the-art for real-time object identification is YOLO V5[4]. For object detection, we can also use Region Based Convolutional Neural Networks (R-CNN) [8]. Crop condition is one of the criteria that goes into making a final decision. Soil condition and weather data are two more variables to consider as shown in Fig. 2.

For collecting meteorological data, we need to use meteorological sensors[5]. Climate and weather are monitored and assessed using meteorological sensors. For identifying the condition of the soil, we can employ an IoT soil condition monitoring sensor. Moreover, we can apply the IoT-Based Soil Condition Monitoring Framework proposed by [9]. We need to feed the data to the AI system for decision-making after acquiring the information utilizing various technologies.

Finally, depending on the criteria, the AI system will make task recommendations using proper machine learning algorithms. This decision-making mechanism will assist farmers in doing suitable farming activities.

---

[4] YOLOv5 | PyTorch
[5] Meteorological Sensors

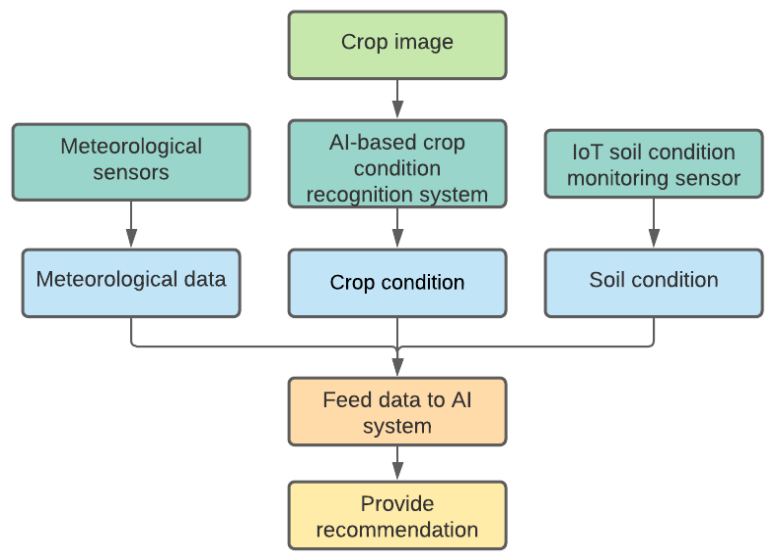

**Fig. 2:** Decision-Making Support System for Agriculture

### 4.2 Water Resource Management System with the help of AI

Water management is needed to achieve SDG #6: Clean Water and Sanitation, SDG #12: Responsible Consumption and Production, and SDG #14: Life under Water. With the help of AI, we propose a water resource management system that will monitor the water quality, forecast the water flow and help to manage according to the situation. A hypothetical framework for smart water management is shown in Fig. 3. The system is responsible for forecasting floods. For measuring water level, different types of sensors[6] can be used. The system can monitor water parameters through water quality sensors[7].

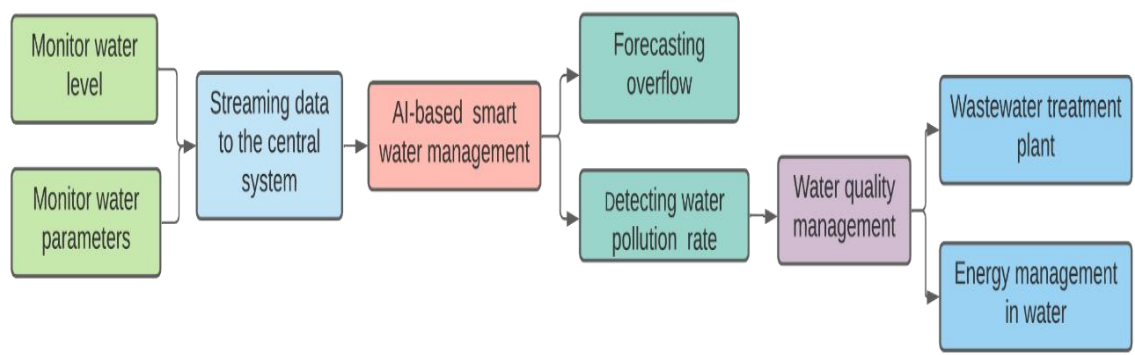

**Fig. 3:** Water Resource Management System

The six main parameters of water quality are bioindicators, dissolved oxygen, turbidity, nitrates, pH scale, and water temperature. Water pollution can be detected using the water quality monitoring framework proposed by [20]. For water quality management, we can utilize WQMCM framework proposed by [21]. Wastewater treatment can be done using the IoT based model proposed by [22]. [23] proposed a smart solution to improve water-energy. We can use the framework for energy management. Based on different indicators, the AI system will forecast and manage the water resource using necessary machine learning algorithms. This system will help predict overflow and measure pollution and manage smartly to achieve sustainability.

---

[6] Water Level Sensors
[7] Water Quality Sensors

## 4.3 Classifying and Managing Waste with AI

Waste management is the most important and largest field in environmental technology. In Bangladesh, waste management is a genuine issue. Most of our waste finds its way to landfills or ocean water. Recycling and reusing are less practiced and since almost all the wastes need to be collected and sorted via human labor, it's time-consuming and not cost-friendly to invest in waste management. Accordingly, an enormous amount of recyclables and reusable wastes are dumped into landfills. Non-biodegradable wastes are overlooked as it takes more human labor to classify and sort. We propose a waste management system that has an easy-to-build infrastructure and provides a cheaper solution than the traditional system.

The inputs of the system are wastes from several sources. For instance, we categorized the sources into three primary regions: household wastes, corporate wastes, and industrial waste. More sources of waste could be added to the system as input, the system function remains the same. After collecting all the wastes from various major sources, it would be gathered at a local waste treatment center. From the local waste treatment center, municipal solid wastes will go through a classification procedure where, with the help of deep learning and robotics, wastes would be classified and sent to a reserved section for a certain type of waste. All the liquid wastes from the local waste treatment center are sent to the liquid and water treatment center. The output of the local waste treatment center for liquid wastes could act as an input for the *Water Resource Management System*.

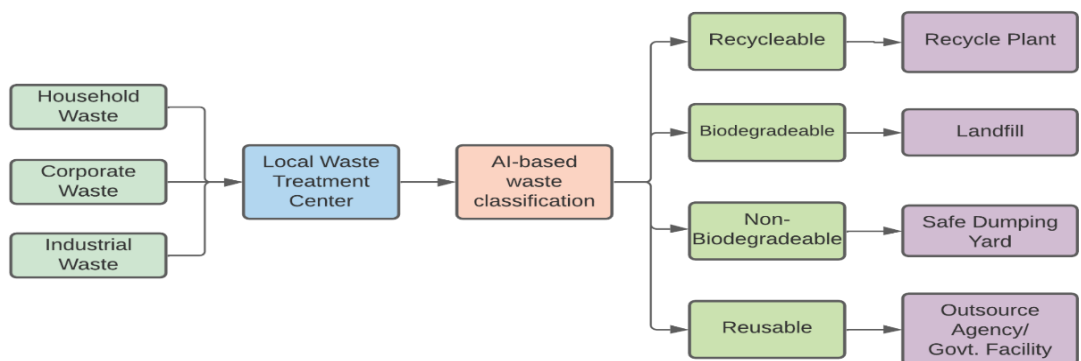

**Fig. 4:** Waste Collection, Treatment and Classification System

Once municipal wastes are classified and sent to a reserved section via a pipeline, a transportation or pipeline system would deliver those to the proper destination. Recyclable wastes will be delivered to recycle plants, biodegradable wastes into landfills, non-biodegradable wastes, or hazardous waste to a safe dumping place, outsource agency or government facilities would collect reusable wastes and treat those to be used again. Convolutional Neural Networks (CNN) or, as discussed in the previous system, YOLO, can be used to classify the wastes. After identifying wastes, mechanical arms or similar robotic mechanisms will sort the waste in a specific order per instruction.

## 4.4 An eco-friendly approach to HVAC systems with the help of AI

HVAC systems are responsible for a large amount of energy consumption and carbon emission. Making the HVAC systems more efficient will help us achieve SDG #11: Sustainable Cities and Communities as well as SDG #13: Climate Action. As the economy of Bangladesh grows and the temperature of cities rise, air conditioning which was once viewed as a luxury is becoming a necessity in order to live in cities like Dhaka. As the number of AC units used in the cities increase, the hot air released by the AC units is having a substantial impact on the temperature surrounding them. As a result, people are buying more AC units to cope with the rising temperature creating a loop of the ever-growing number of AC units in the cities. In order to make sure the cities of Bangladesh doesn't reach an over air-conditioning situation like HongKong[8], we need to build centralized HVAC systems that take both efficiency and environmental effects into account. Here, we propose an eco-friendly approach to a centralized HVAC system with the use of AI.

In order to make an efficient HVAC system that also tries to minimize its effect on the surrounding environment, we will need to monitor both indoor and outdoor environmental parameters like temperature, humidity and air purity using air quality sensors. The monitored data will be sent to an AI model which will in turn predict future indoor and outdoor environmental parameters based on historical data. We will use an optimum demand response scheduling with the help of methods proposed by Kim et al. [16] & Wang et al. [17] We will then use another AI model to predict the effects of the HVAC system on the outside environment. Finally, we will feed the prediction data of the two models to another AI model that will suggest an optimal HVAC usage pattern that has the least amount of environmental impact.

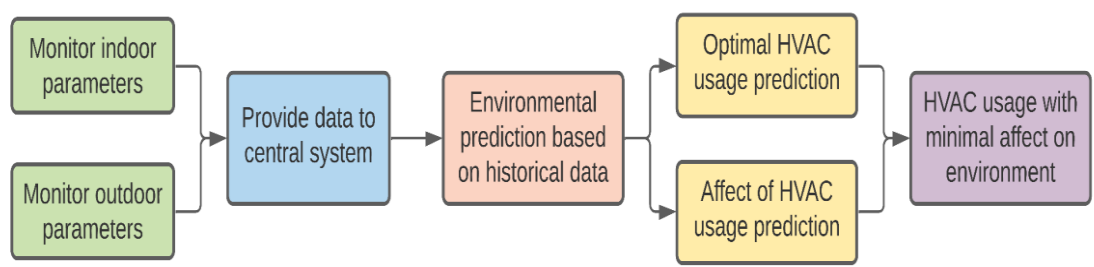

**Fig. 5:** Finding optimal HVAC usage pattern with minimal effect on the surrounding environment

## 5. Negative Impact of AI

AI is a vast computer science subject that focuses on developing intelligent machines that can do tasks that would otherwise need human intellect. As a result, AI is a frequently utilized tool for attaining long-term progress. However, this technique has several disadvantages. In this part, we'll discuss AI's detrimental influence on sustainable development.

AI has the potential to exacerbate environmental deterioration, according to the forum and experts on the subject [10]. The usage of high-power GPUs for machine learning training has previously been linked to an increase in $CO_2$ emissions which is the major cause of global warming. Because the environment, economy, and society are all intertwined, global warming has the potential to harm not just the environment, but also the way of life of living creatures. To achieve sustainable development, any of the pillars of sustainable development is essential. Another big disadvantage of AI technology is that it generates a lot of e-waste, which is extremely difficult to control using IWM (Integrated Waste Management) techniques. AI training is an energy-intensive process. New estimates indicate that carbon dioxide emissions when training a single AI are up to 284 tonnes of carbon dioxide. This is five times the lifetime emissions of an average car [11].

The research and deployment of AI are in the control of huge corporations. This is due to the fact that AI requires cash, and society's tremendous wealth is typically concentrated in the hands of a few individuals. We are becoming sedentary as a result of the increased usage of AI-based technology, which is contributing to our health problems. The substitution effect of AI will not only increase unemployment but will also reduce wages, resulting in lower income for low-end workers, exacerbating the social divide between rich and poor.

---

[8] How did Hong Kong become addicted to air conditioning?

## 6. Discussion

This research focuses on achieving SDGs with the help of AI and modern automation technology. Our goal is to replace traditional inefficient systems with efficient and eco-friendly ones. Although AI and modern technologies are being used daily to make a traditional system more efficient in every sector, the use of AI in environmental development is overlooked even to date. We found that most of the research based on AI on sustainable development addresses the statistical problems we face to maintain sustainable development. While these papers proposed and presented smart solutions in terms of sustainable development, most of these researches are tunnel-visioned into a single category.

In this paper, we proposed a framework that includes agriculture, waste management, clean water, and power-efficient cooling solutions for households and industry. Our framework proposes to replace traditional systems that require a large amount of human labor, are inefficient, and create a negative impact on the environment. The proposed framework provides a replacement solution while being more efficient in carbon emission, energy usage, environmental pollution, and climate change.

While our solutions provide better and more efficient results in the paper, there are some challenges to be addressed in the context of Bangladesh or any other under-developed country. Primarily, it is difficult to take crop images using drones. It can increase the cost and drones are not permitted everywhere. Our Waste Management System and Water Resource Management System is relatively expensive to develop for a country like Bangladesh, but it is mandatory for us to take steps to manage waste properly to grow towards sustainable development. Finally, our HVAC (Heating Ventilation Air Conditioning) system is going to have a great impact on climate change considering it consumes less energy, less power and produces less heat output, and provides efficient cooling solutions both for residential and industrial scenarios. It will be costly to integrate with old hardware and it needs a centralized control center, to begin with. Besides, the machine learning models have to be area-specific, meaning we need to build a control center in each area to achieve the best results. This system might seem irrelevant at first, but with the rapid growth in urbanization and increasing amount of data centers, we must shift to an efficient system like the one we presupposed to maintain the sustainable development goals.

We hope to improve the framework in the future with more data and a wide range of integrated systems. To implement the systems in a real-world scenario, we need to make sure that all the stakeholders are informed about the use of AI and our systems co-operates with the rules and regulations of the area. We have to raise awareness and educate every stakeholder of our systems to achieve optimal output.

## 7. Conclusion

The coming era of AI will change many aspects of our day-to-day lives. From the assistant in our phones to the Self-Reliant Rovers [18] on mars, AI has already impacted our lives. It is high time we used AI in major aspects of our lives to extract the most from AI technologies to make our lives more efficient and sustainable. In this paper, we have explored the role of AI in sustainable development and how AI-based frameworks can be used for achieving sustainable development goals in developing countries like Bangladesh. Four case studies were made on some of the major sustainable development sectors. These studies take a look into the usage and research of AI in major fields of sustainable development, like agriculture, water management, waste management, and HVAC systems, to highlight the potential of AI in sustainable development. Based on these extensive case studies on major fields of sustainable development, we proposed an AI-based framework for developing countries like Bangladesh that can be used to achieve major sustainable development goals.

To conclude, we discuss the challenges of implementing such frameworks and how the negative impact can be mitigated are discussed in detail. The research shows that if AI technologies are used with good intent and proper rules and regulations are followed, the positive impact of AI far surpasses the negative impact of AI. So, proper rules and regulations should be made to regulate AI usage while encouraging people to use AI in their day-to-day life to make our lives more sustainable.

**References**

[1] Keeble, B.R., 1988. The Brundtland report:'Our common future'. Medicine and war, 4(1), pp.17-25.

[2] Shamsuzzaman, M.M. and Islam, M.M., 2018. Analysing the legal framework of marine living resources management in Bangladesh: Towards achieving Sustainable Development Goal 14. *Marine Policy*, *87*, pp.255-262.

[3] Islam, M.M. and Shamsuddoha, M.D., 2018. Coastal and marine conservation strategy for Bangladesh in the context of achieving blue growth and sustainable development goals (SDGs). *Environmental science & policy*, 87, pp.45-54.

[4] Sourav, A.I., Lynn, N.D. and Santoso, A.J., 2020, November. Designing a conceptual framework of a smart city for sustainable development in Bangladesh. In *Journal of Physics: Conference Series* (Vol. 1641, No. 1, p. 012112). IOP Publishing.

[5] Roy, J., Islam, S.T. and Pal, I., 2021. Implementation framework for sustainable development: What matters in the context of Bangladesh. *International Energy Journal*, *21*(1A).

[6] Chostner, B., 2017. See & Spray: the next generation of weed control. *Resource Magazine*, 24(4), pp.4-5.

[7] Redmon, J., Divvala, S., Girshick, R. and Farhadi, A., 2016. You only look once: Unified, real-time object detection. In *Proceedings of the IEEE conference on computer vision and pattern recognition* (pp. 779-788).

[8] Girshick, R., Donahue, J., Darrell, T. and Malik, J., 2015. Region-based convolutional networks for accurate object detection and segmentation. *IEEE transactions on pattern analysis and machine intelligence*, 38(1), pp.142-158.

[9] Manickam, S., 2020. IoT-Based Soil Condition Monitoring Framework.

[10] Herweijer, C. and Waughray, D., 2018. Harnessing artificial intelligence for the earth. *Fourth Industrial Revolution for the Earth Series*.

[11] Yudkowsky, E., 2008. Artificial intelligence as a positive and negative factor in global risk. Global catastrophic risks, 1(303), p.184.

[12] CleanRobotics. (n.d.). TrashBot: a smart waste bin by CleanRobotics. Available at: https://cleanrobotics.com/trashbot/.

[13] Mitchell, T. (1997) "Machine learning". McGraw hill New York

[14] LeCun, Y., Bengio, Y. en Hinton, G. (2015) "Deep learning", *nature*. Nature Publishing Group, 521(7553), bll 436–444.

[15] Evans, R. en Gao, J. (2016) "Deepmind AI reduces Google data center cooling bill by 40%",

*DeepMind blog*, 20, bl 158.

[16] Kim, Y.-J. (2020) "A supervised-learning-based strategy for optimal demand response of an HVAC system in a multi-zone office building", *IEEE Transactions on Smart Grid*. IEEE, 11(5), bll 4212–4226.

[17] Wang, J. *et al.* (2020) "Operating a commercial building HVAC load as a virtual battery through airflow control", *IEEE Transactions on Sustainable Energy*. IEEE, 12(1), bll 158–168.

[18] Gaines, D., Russino, J., Doran, G., Mackey, R., Paton, M., Rothrock, B., Schaffer, S., Agha-Mohammadi, A.A., Joswig, C., Justice, H. and Kolcio, K., 2018. Self-Reliant Rover Design for Increasing Mission Productivity.

[19] Caiado, J. (2007). Forecasting water consumption in Spain using univariate time series models.

[20] Altenburger, R., Brack, W., Burgess, R. M., Busch, W., Escher, B. I., Focks, A., et al.(2019). Future water quality monitoring: Improving the balance between exposure and toxicity assessments of real-world pollutant mixtures. *Environmental Sciences Europe*,31(1). doi: 10.1186/s12302-019-0193-

[21] Li, D., & Liu, S. (2019). Wireless sensor networks in water quality monitoring. *Water Quality Monitoring and Management*, 55–100. doi: 10.1016/b978-0-12-811330-1.00002-8

[22] Pasika, S., & Gandla, S. T. (2020). Smart water quality monitoring system with cost-effective using iot.Heliyon,6(7). doi: 10.1016/j.heliyon.2020.e0409

[23] Helmbrecht, J., Pastor, J., & Moya, C. (2017). Smart solution to improve water-energy nexus for Water Supply Systems. Procedia Engineering, 186, 10